\documentclass[conference]{IEEEtran}

\usepackage{epsfig}
\usepackage{graphicx}
\usepackage{amsmath}
\usepackage{amssymb}
\usepackage{booktabs}
\usepackage{rotating}
\usepackage{array}
\renewcommand{\arraystretch}{1.2}

\def\eg{\emph{e.g. }}

\def\wrt{w.r.t. } 
\def\etal{\emph{et al. }}

\ifCLASSOPTIONcompsoc
  \usepackage[nocompress]{cite}
\else
  \usepackage{cite}
\fi

\begin{document}
\title{
Rapid Light Field Depth Estimation with Semi-Global Matching 
}
\iffalse
\author{
\IEEEauthorblockN{ICCP 2019 Anonymous submission \# 31\\\\\\\\\\}
}
\else
\author{
\IEEEauthorblockN{Yuriy Anisimov, Oliver Wasenm\"uller, Didier Stricker}
\IEEEauthorblockA{Department Augmented Vision, German Research Center for Artificial Intelligence (DFKI)\\
Department of Computer Science, University of Kaiserslautern\\
Kaiserslautern, Germany\\
\{yuriy.anisimov, oliver.wasenmueller, didier.stricker\}@dfki.de}}
\fi
\iftrue
\makeatletter
\def\ps@IEEEtitlepagestyle{
  \def\@oddfoot{\mycopyrightnotice}
  \def\@evenfoot{}
}
\def\mycopyrightnotice{
  {\footnotesize 978-1-7281-4914-1/19/\$31.00~\copyright~2019 IEEE\hfill} 
  \gdef\mycopyrightnotice{}
}
\fi
\maketitle

\begin{abstract}
Running time of the light field depth estimation algorithms is typically high. 
This assessment is based on the computational complexity of existing methods and the large amounts of data involved.
The aim of our work is to develop a simple and fast algorithm for accurate depth computation.
In this context, we propose an approach, which involves Semi-Global Matching for the processing of light field images.
It forms on comparison of pixels' correspondences with different metrics in the substantially bounded light field space.
We show that our method is suitable for the fast production of a proper result in a variety of light field configurations.
\par
\end{abstract}

\section{Introduction}
\label{sec:intro}
In terms of computer vision a light field, originally described by Gershun \cite{gershun1939light}, can be interpreted as a set of 2-dimensional images, projected from a scene with an equal physical distance (baseline) between the adjacent viewpoints in the horizontal or vertical direction.
\par
Real-world capturing of such a set can be done in different ways.
The simplest one involves an ordinary camera on the moving stage, which is shifted on constant length for every viewpoint, producing a 3-dimensional light field.
Light fields can be captured by so-called plenoptic camera \cite{lumsdaine2009focused}, which has a micro-lens array placed in front of the sensor image plane, providing a 4-dimensional light field.
Such cameras were presented by Ng \etal \cite{ng2005light} and Perwass \etal \cite{perwass2012single}.
Multi-camera arrays with individual lenses can be also used for light field capturing.
A large-scale version of this camera type was proposed by Wiburn \etal \cite{wiburn2004high}, whereas small-scale versions were presented by Venkataraman \etal \cite{venkataraman2013picam} and Anisimov \etal \cite{anisimov2019compact}
\par
Light field cameras are attracting considerable interest due to the possibility of their utilization in different industrial (optical inspection), biological (three-dimensional microscopy) and cinematic applications.
This demand can be explained by the features of light field images, \eg digital refocusing, which is an ability to change the focus on already acquired images \cite{ng2005light}.
Another useful feature of light field images is a possibility of accurate depth estimation. 
In contrast to classical multi-view systems, the matching correspondence search can be simplified by preserving the constant baseline between views in light field images.
\par
\begin{figure}
\begin{center}
\begin{tabular}{cccc}
\includegraphics[height=32.5mm]{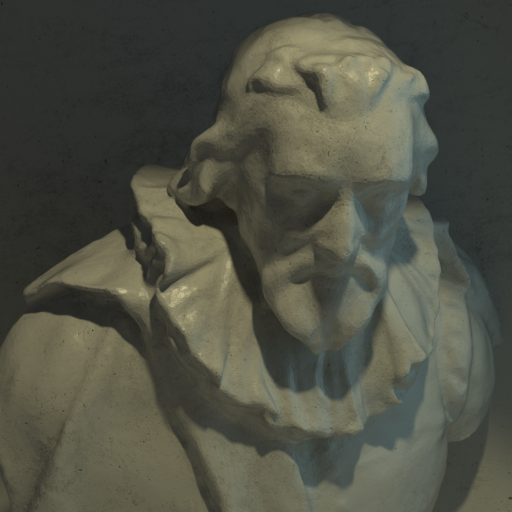} &\phantom{abc} \includegraphics[height=32.5mm]{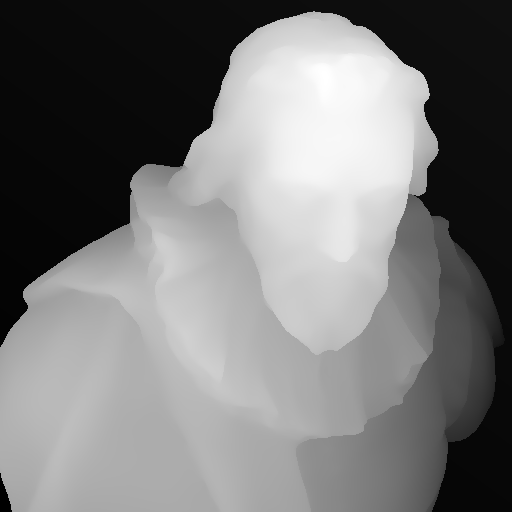} \\ (a) & (b) \\[6pt]
\includegraphics[height=32.5mm]{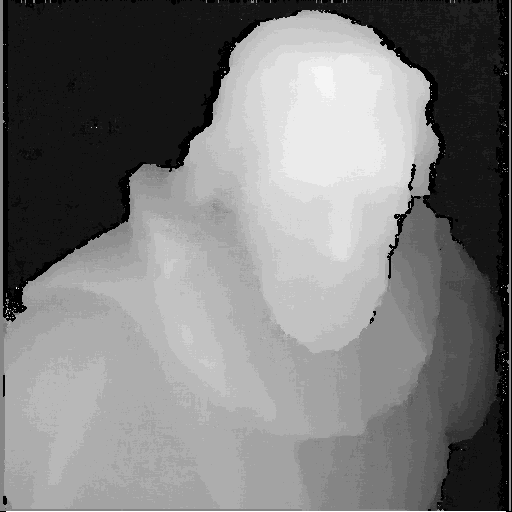} & \phantom{abc}\includegraphics[height=32.5mm]{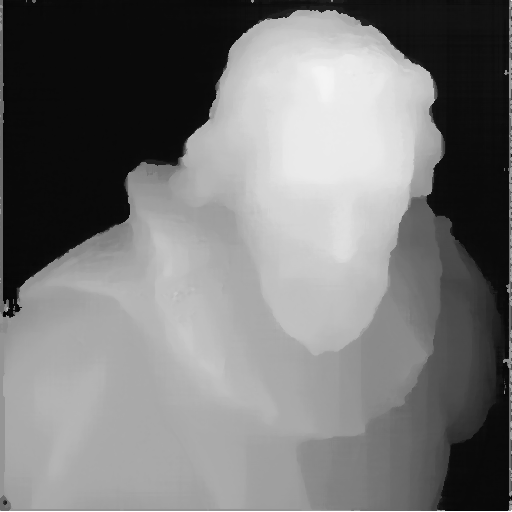} \\ (c) & (d) \\[6pt]
\end{tabular}
\caption{Result of our algorithm for a synthetic scene from 4D Light Field Benchmark  \cite{4DLFB, honauer2016dataset}: (a) Center image from a light field, (b) ground truth, (c) initial disparity map, (d) final disparity map.}
\label{fig:rw_scenes}
\end{center}
\vspace{-8mm}
\end{figure}
An overview of the actual light field depth estimation algorithms is provided in Section \ref{light_field_processing_sota}.
We can state a major challenge in this area as a trade-off between runtime and the depth quality.
Existing approaches are computationally demanding and involve large amounts of data for processing.
It limits rapid depth map calculation.
Most studies have tended to focus on depth map accuracy rather than execution time.
We mainly address the runtime issue and aim towards real-time processing.
\par
The proposed algorithm is based on pixel matching in light field space with different similarity measurements for estimation of a dense depth map. 
We choose Census and $\ell_{2}$-norm for generation of matching cost, which is explained in sections \ref{image_similarity_measurement} and \ref{matching_cost_generation}. 
Although this principle gives good results in terms of depth map quality, computational demand and consequential running time of this method do not allow its direct use for fast or even real-time estimation.
In order to reduce the number of calculations for this step, we generate the initial disparity map from some of the views in light field image and use this information as "boundaries" for high-intensive correspondence search among all views, which is outlined in Section \ref{reducing_of_sampled_disparity_hypotheses}.
For disparity quality improvement a semi-global matching (SGM) method \cite{hirschmuller2005accurate} is applied for previously generated costs. 
This method recommended itself as one of the most developed and optimized for providing accurate real-time depth results, which is fully justified by works reviewed in Section \ref{semi_global_matching_sota}.
\par
%
The presented approach follows the core idea of Anisimov \etal \cite{anisimov2018fast, anisimov2019compact}, but introduces the following features.
We consider our work to be the first attempt to apply the SGM method directly to matching costs from light field images. 
For initial disparity map calculation we use a fusion of disparity maps for different views \wrt reference view and do not apply refinement to it, unlike in \cite{anisimov2018fast}.
Together with that, we propose the utilization of Census transform for RGB images, in contrast to classical grayscaled images. 
As a result of all these contributions, we overcome limitations of previous algorithm \cite{anisimov2018fast} such as objects' boundaries sharpness and noise in discontinuity areas.
The workflow of our method is presented in Section \ref{proposed_approach}.
\par
This approach achieves almost the best running time among other methods together with acceptable depth quality level.
We provide a verification of our results by 4-dimensional Light Field Benchmark \cite{4DLFB, honauer2016dataset} in Section \ref{experiments}.
Together with that, we present real-world qualitative results.
On top of the benchmark estimators, we use a M-metric, proposed in \cite{anisimov2018fast}, to evaluate algorithms in terms of efficiently estimated pixels per unit of time.
Also, we show the usage of different image similarity measurements methods for matching cost estimation dependently of various light field configuration properties.
\section{Related work} \label{related_work}
\subsection{Light field processing} \label{light_field_processing_sota}
Several light field depth estimation approaches utilize Epipolar Plane Image (EPI) structure, proposed by Bolles \etal \cite{bolles1987epipolar} and defined as a "slice" of the light field, in which the slope of the line with reference to a certain point in an image is proportional to the depth value.
Wanner and Goldluecke present a reformulation of stereo matching to a constrained labeling problem on EPIs with further variational regularization \cite{wanner2012globally}.
Kim \etal \cite{kim2013scene} present an approach for depth reconstruction from high-resolution images using a slope analysis for the lines in EPIs.
Wang \etal \cite{wang2015occlusion} extend this approach for the occlusions-handling case.
Johannsen \etal \cite{johannsen2016sparse} present a method based on a sparse decomposition of a light field with further depth-orientation dependency retrieving.
Sheng \etal \cite{sheng2018occlusion} extract multi-orientation EPIs and aggregate local depth maps from them with the preservation of occlusions.
Several methods do not consider EPI for depth estimation.
Neri \etal \cite{neri2015multi} adaptively combine data term and multi-view stereo with a multi-resolution approach for reducing the complexity of the algorithm.
Sabater \etal \cite{sabater2017dataset} propose a method for real-time depth estimation, using a zero-normalized cross-correlation as an image similarity measurement together with pyramid strategy for coarse-to-fine reconstruction.
An approach from Jeon \etal \cite{jeon2015accurate} utilizes the cost volume from shifted sub-aperture images with different similarity measurements and further result refinement via Graph Cuts (GC) \cite{kolmogorov2002multi}.
Huang \cite{huang2017robust} presents a novel framework derived from Markov random fields \cite{blake2011markov}.
Based on this framework, an empirical Bayesian algorithm for depth estimation is developed.
Anisimov and Stricker \cite{anisimov2018fast} propose an efficient approach for depth estimation by initializing the line fitting hypothesis search with a result of SGM.
In this work we follow this core idea, but introduce some unique features as mentioned in Section \ref{sec:intro}.
\par
In recent years interest in the usage of neural networks for the light field processing has been growing.
These methods can produce a relatively fast depth result on high-performance graphics processing units (GPU).
Heber and Pock \cite{heber2016convolutional} utilize a Convolutional Neural Network (CNN) for predicting the orientation of a 2-dimensional hyperplane, which represent the depth information, in the 4-dimensional light field space.
Sun \etal \cite{sun2016data} augment the EPI with Hough and Radon transform, providing the modified result to CNN.
Jeon \etal \cite{jeon2018depth} present an extension of the algorithm \cite{jeon2015accurate} with learning-based matching costs.
A paper by Shin \etal \cite{shin2018epinet} explains an architecture of the multi-stream network, which currently achieves the top result in 4D Light Field Benchmark \cite{4DLFB, honauer2016dataset}.
\subsection{Semi-global matching} \label{semi_global_matching_sota}
Original SGM description was proposed by Hirschmuller \cite{hirschmuller2005accurate}.
Because of the high-quality output and relatively small runtime, this algorithm found its application in stereo reconstruction-related fields.
Our method utilizes the SGM routine described by Haller \etal \cite{haller2010real} with the cost aggregation principles from \cite{hirschmuller2005accurate}.
Recent work of Hernandez-Juarez \etal \cite{sgm_gpu_iccs2016} shows outstanding real-time results of SGM on embedded GPU.
They present a parallel version of this approach with Center-Symmetric Census transform as a cost generation method.
One approach with the SGM extension for an arbitrary number of images is presented by Bethmann and Luhmann \cite{bethmann2015semi}.
Matching cost values in their method are computed as voxels in the object space, in which the minimization process is performed.
\par
\section{Proposed approach} \label{proposed_approach}
This section provides an overview of the steps of the proposed algorithm.
We describe light field parametrization in Section \ref{light_field_parametrization} and provide a comparison of image similarity metrics in Section \ref{image_similarity_measurement}.
Estimation of matching cost among the views in light field images is described in Section \ref{matching_cost_generation}.
Forming of initial disparity map is presented in Section \ref{reducing_of_sampled_disparity_hypotheses} and the final result generation is outlined in Section \ref{extended_semi_global_matching}.
Result post-processing is described in Section \ref{final_steps}.
\par
\subsection{Light field parametrization} \label{light_field_parametrization}
Levoy and Hanrahan \cite{levoy1996light} describe a light field in the form of two-plane parametrization: a plane of spatial coordinates $(u, v)$ for a 2-dimensional view in the light field image, and angular plane $(s, t)$ for the viewpoint representation.
The whole light field space can be denoted as $L(u, v, s, t)$.
The way of finding the matching pixel for a specified disparity hypothesis in light field views can be formulated using this definition. 
For a given reference light field view $(\hat s, \hat t)$ and a disparity hypothesis $d$, matching pixel position $(u, v)$ in the view $(s, t)$ can be determined as \cite{kim2013scene}: 
\begin{align}
\hat{p}(u, v, s, t, d) = L(u + (\hat s - s)d, v + (\hat t - t)d, s, t).
\label{align:pix_match}
\end{align}
\subsection{Image similarity measurement} \label{image_similarity_measurement}
Generation of a matching cost, based on the comparison of image elements such as pixels, patches or windows, is an important step for disparity estimation algorithms.
Commonly used pixel-wise functions are "city block" and Euclidean distance \cite{chen2005similarity} ($\ell_{1}$- and $\ell_{2}$-norm respectively).
The result of these functions can be used directly for the cost forming or can be further improved in a form of Earth Mover Distance \cite{rubner2000earth} or Kernel Density Estimation (KDE) \cite{parzen1962estimation}.
Algorithms from \cite{kim2013scene} and \cite{anisimov2018fast} are using KDE in a form of Epanechnikov kernel \cite{epanechnikov1969non}.
\par
In window-based methods, widely used measures are the sum of absolute differences, the sum of squared differences and normalized cross-correlation \cite{hirschmuller2009evaluation}.
These estimators can provide more accurate results in contrast to pixel-based methods, but the computation time increases since for each pixel in an image more pixels around are involved, which can limit usage on window-based approaches in rapid estimation algorithms. 
\par
One of the efficiently used method for the matching cost generation is Census transform as detailed by Zabih \etal \cite{zabih1994non}.
Radiance value of a pixel in an image $I$ is compared to pixels nearby within a set of pixels coordinates $D$, lying in a window.
It results in a bit string for a pixel in Census-transformed image $I_{c}$: 
\begin{align}
I_{c}(u,v)=\underset{[i,j]\in D}{\bigotimes}\xi(I(u, v),\,I(u+i, v+j)),
\end{align}
where $\otimes$ stands for bit-wise concatenation.
Pixel relations are defined as: 
\begin{align}
{\xi(p_1, p_2)} =
\begin{cases}
0, &p_1 \leqslant p_2\\
1,& p_1 > p_2
\end{cases}
.
\end{align}
Although this method is based on window around a pixel, the number of calculations can be reduced by using a sparse window for census transform. 
\subsection{Matching cost generation} \label{matching_cost_generation}
Two image similarity measurements are used in this work for the generation of matching cost. 
A Census-based matching cost function can be defined through a Hamming distance between corresponding pixels from Census-transformed images.
For two images in Census-transformed light field $L_{c}$ with coordinates $(\hat s, \hat t)$ and $(s, t)$: 
\begin{align}
C_{c}(u, v, d) = HD(L_{c}(u, v, \hat s, \hat t), \hat{p}_{c}(u, v, s, t, d)),
\label{align:cost_census}
\end{align}
where $\hat{p}_{c}$ stands for matching pixel position in $L_{c}$ (Eq. \ref{align:pix_match}).
$HD$ is the Hamming distance function.
For two vectors $x_i$ and $x_j$ ($| x_i | = | x_j | = n$, here and further $| \ldots |$ denotes cardinality) it can be determined as a number of elements with different values.
\begin{align}
HD(x_i, x_j) = \sum\limits_{k=1}^n x_{ik} \oplus x_{jk}.
\label{align:hamming_distance}
\end{align}
In our approach Census transformation is extended to be applied for RGB images.
It is performed for each view channel separately, and the result is composed by a sum of Hamming distances of pixels in every channel.
Cost, generated by this method, is used in calculations of initial disparity map, explained in Section \ref{reducing_of_sampled_disparity_hypotheses}.
\par
Radiance comparison for correspondence matching in light field space is performed by $\ell2$-norm. 
For two light field views with coordinates $(\hat s, \hat t)$ and $(s, t)$ is determined as: 
\begin{align}
C_{\ell2}(u, v, d) = \|L(u, v, \hat s, \hat t) - \hat{p}(u, v, s, t, d)\|_{2}.
\label{align:euclidean_distance}
\end{align}
Cost from this method is used during the final disparity map estimation, described in Section \ref{extended_semi_global_matching}. 
\subsection{Initial disparity map} \label{reducing_of_sampled_disparity_hypotheses}
\newcommand{\ra}[1]{\renewcommand{\arraystretch}{#1}}
\begin{table*}[]\centering
\ra{1.3}
\begin{center}
\begin{tabular}{lrrrrrrrrrrr}\toprule

         & \multicolumn{2}{c}{$BadPix$} &\phantom{abc}& \multicolumn{2}{c}{$MSE$} &\phantom{abc}& \multicolumn{2}{c}{$Runtime$, $\log_{10}$ } &\phantom{abc}& \multicolumn{2}{c}{$M$, \%/s} \\
           \cmidrule{2-3} \cmidrule{5-6} \cmidrule{8-9} \cmidrule{11-12}
         & Median & Average && Median & Average && Median & Average && Median & Average       \\
\midrule
BSL \cite{anisimov2018fast}      & 13.41                      & 12.74                       &                      & 5.43                       & 7.28                        &                      & 0.71                       & 0.78                        &                      & 18.39                      & 22.25                       \\
EPI1 \cite{johannsen2016sparse}  & 22.89                      & 24.32                       &                      & 3.93                       & 5.98                        &                      & 1.93                       & 1.95                        &                      & 0.91                       & 0.87                        \\
EPI2 \cite{wanner2012globally}   & 22.94                      & 22.65                       &                      & 5.72                       & 8.24                        &                      & 0.94                       & 0.92                        &                      & 9.05                       & 9.31                        \\
EPINET \cite{shin2018epinet}     & \textbf{3.38}                       & \textbf{4.93}                        &                      & \textbf{1.21}                       & \textbf{2.48}                        &                      & 0.29                       & 0.30                        &                      & \textbf{48.91}                      & 48.12                       \\
LF \cite{jeon2015accurate}       & 16.15                      & 16.19                       &                      & 7.96                       & 9.13                        &                      & 3.00                       & 3.00                        &                      & 0.08                       & 0.08                        \\
LF\_OCC \cite{wang2015occlusion} & 17.82                      & 15.07                       &                      & 2.70                       & 6.76                        &                      & 2.69                       & 2.72                        &                      & 0.17                       & 0.17                        \\
OFSY \cite{strecke2017accurate}  & 11.33                      & 12.04                       &                      & 5.43                       & 7.03                        &                      & 2.30                       & 2.30                        &                      & 0.46                       & 0.47                        \\
RM3DE \cite{neri2015multi}       & 7.99                       & 10.22                       &                      & 1.46                       & 3.92                        &                      & 1.65                       & 1.68                        &                      & 1.96                       & 1.95                        \\
RPRF \cite{huang2017robust}      & 9.89                       & 10.02                       &                      & 3.76                       & 5.68                        &                      & 1.55                       & 1.54                        &                      & 2.53                       & 2.64                        \\
FSL (ours)                        & 11.92                       & 12.95                        &                      & 3.97                       & 6.64                        &                      & \textbf{0.25}                      & \textbf{0.23}                      &                      & 48.33                    & \textbf{56.09}          \\          
\bottomrule
\end{tabular}
\end{center}
\caption{Evaluation of different algorithms with general metrics on 4D Light Field Benchmark \cite{4DLFB, honauer2016dataset}}
\label{table:4dlfb}
\vspace{-2mm}
\end{table*}
In order to create boundary values for correspondence search in whole light field space by more computationally-intensive algorithm the initial disparity map is calculated by using lower computationally-intensive algorithm.  
According to this concept, we first estimate the disparity maps for the set of four cross-lying views in the light field image $V = \{(\hat s, 0),(\hat s, t_{max}), (0, \hat t), (s_{max}, \hat t)\}$, where $s_{max}$ and $t_{max}$ correspond to horizontal and vertical angular dimensions of light field image. 
Disparity maps are estimated relative to reference view $(\hat s, \hat t)$. 
Calculations in this step are simplified compared to whole light field space correspondence matching by reducing the number of processed views and preserving changes only in one angular direction. 
Corresponding Census-transformed views from a light field image are used for this estimation. For every view in a set $V$ the cost is achieved as 
\begin{align}
C_{c_{i}}(u, v, d) = HD(L_{c}(u, v, \hat s, \hat t), \hat{p_{c}}(u, v, V_{i}, d)),
\label{align:initial_disparity_cost}
\end{align}
where $i = [1,|V|]$, and disparity hypothesis $d$ lies in predefined range $d \in T = [d_{min}, d_{max}]$, which covers all possible shifts of the pixels within two opposing views on one angular axis \eg $(\hat s, 0)$ and $(\hat s, t_{max})$. 
\par
Results of the cost matching are then individually aggregated with original SGM method.
For each pixel $p = (u, v)$ and $d \in T$, after traversing in direction $r$, formulated as a 2-dimensional vector $r$ = \{$\Delta u$, $\Delta v$\}, aggregated cost $L_r$ is  
\begin{align}
\begin{split}
&L_r(p,d) = C(p, d)+\\
&\min\,(L_r(p - r, d),\\
&L_r(p - r, d - 1) + P\mathit{1},\\
&L_r(p - r, d + 1) + P\mathit{1},\\
&\underset{t}{\min}\,L_r(p - r, t) + P\mathit{2}),
\end{split}
\label{align:sgm}
\end{align}
where $P\mathit{1}$ and $P\mathit{2}$ are penalty parameters for neighbourhood disparities, $P\mathit{2} \geqslant P\mathit{1}$ and $C(p, d) = C_{c}(u, v, d)$ for the case described in this section.
Costs are then summarized among all directions:
\begin{align}
C_s(p,d) = \underset{r}{\sum}\,L_r(p,d).
\label{align:sum}
\end{align} 
We compute disparity map separately using the winner-takes-all (WTA) strategy on the summarized cost: 
\begin{align}
D_s(p) =\underset{d}{\arg\min}\, C_s(p, d).
\label{align:wta}
\end{align}
As a result we obtain four intermediate disparity maps $D_{i}, i = 1 ... |V|$, which already correspond to the reference view of light field image, unlike in the algorithm of Anisimov and Stricker \cite{anisimov2018fast}, where disparity maps had to be projected on the coordinates of the reference view. 
For obtaining the initial disparity map $D_{init}$ the intermediate maps are fused using a confidence threshold $\varphi$. 
As an initialization step $D_{init} = D_{1}$, for every pixel $(u,v)$ we verify if inequality $|D_{init}(u,v) - D{i}(u,v)| < \varphi$ for the rest of intermediate maps is true. 
If it is so -- new value in initial disparity map is defined as an average between $D_{init}(u,v)$ and $D_{i}(u,v)$, if not -- the pixel is discarded as uncertain. 
To partially cover "holes" without a valid disparity value, which appear after the described fusion, we apply a one- or two-pass median-based filling of particularly this holes with nearby values within a window.
An example of initial disparity map can be found in Fig. \ref{fig:rw_scenes} (c). 
\par
$D_{init}$ is used for generation of boundaries for the further estimation. 
These boundaries will limit matching cost generation in the whole light field space. 
Two structures named high and low borders ($D_{H}$ and $D_{L}$ respectively) are generated by using the border threshold $\lambda$ in such a manner:
\begin{align}
D_{H}(u,v) = D_{init}(u,v) + \lambda; D_{L}(u,v) = D_{init}(u,v) - \lambda.
\label{align:borders}
\end{align}
The values, which lies outside on predefined disparity range ($D_{H} > d_{max}$, $D_{L} < d_{min}$) are saturated accordingly. Invalid values from $D_{init}$ are marked in the corresponding borders for re-computation on the whole disparity range $T$. Also, we exclude edges from the borders, so the objects boundaries are reconstructed better by involving all information from light field space. For that, the technique with Sobel operator \cite{sobel19683x3} for gradient extraction from reference view is applied.
\par
\subsection{Final disparity map} \label{extended_semi_global_matching}
Within the border for disparity values, we perform a correspondence search across all views in light field image for computing matching cost.
Matching cost $S$ for a pixel $p = (u, v)$  with respect to the reference view $(\hat s, \hat t)$ for each possible hypothesis $d \in [D_{L}(p),D_{H}(p)]$ can be computed using the $\ell_{2}$-norm as 
\begin{align}
\begin{split}
S(u, v, d) = {\sum\limits_{s=1}^n\sum\limits_{t=1}^m}{\|L(u, v, \hat s, \hat t) - \hat{p}(u, v, s, t, d)\|_{2}},
\label{align:S_L2}
\end{split}
\end{align}
We do not use generated cost values directly for disparity estimation. 
Instead, for the result improvement, SGM method is used to aggregate the generated matching cost.
Eq. \ref{align:sgm} is applied to the $S(u,v,d)$ instead of $C_{s}(p,d)$, result for every pixel traversing direction is provided to \label{align:sum}
$C_{s_\ell2}$. 
SGM in this step performed only for the disparity values, which are in the predefined bounded range $d \in [D_{L}(p),D_{H}(p)]$, so values outside the boundaries do not affect the aggregation process.
We determine the final disparity map $D_{f}$ according to the lowest value of the aggregated costs $C_{s_\ell2}$ for each pixel by applying the WTA strategy for Eq. \ref{align:wta}.
\par
\begin{figure*}
\begin{center}
\centering
\setlength{\tabcolsep}{2pt}
\begin{tabular}{cccccc}
\includegraphics[height=27.5mm]{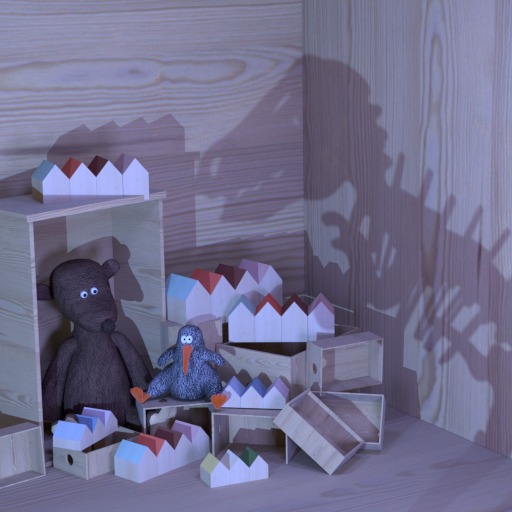} & \includegraphics[height=27.5mm]{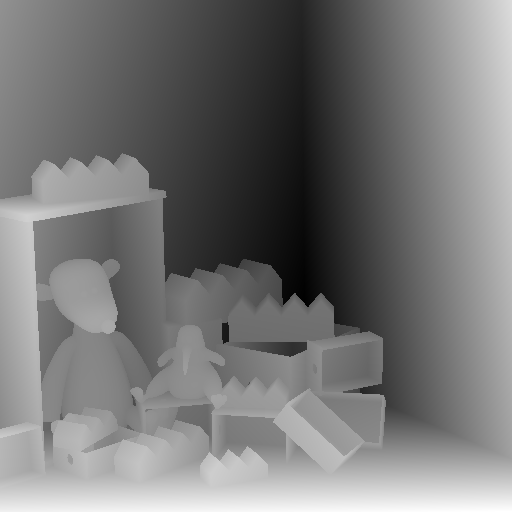} & \includegraphics[height=27.5mm]{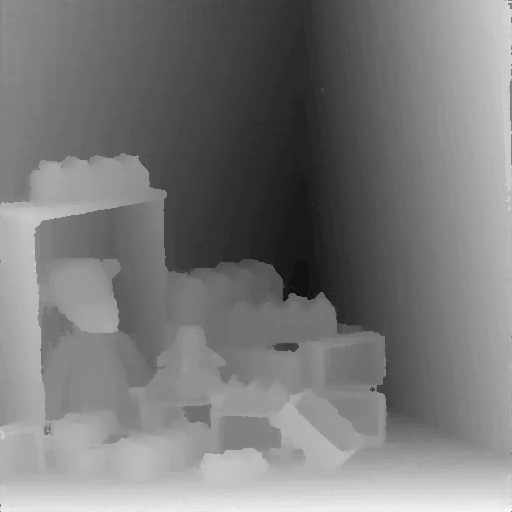} &  \includegraphics[height=27.5mm]{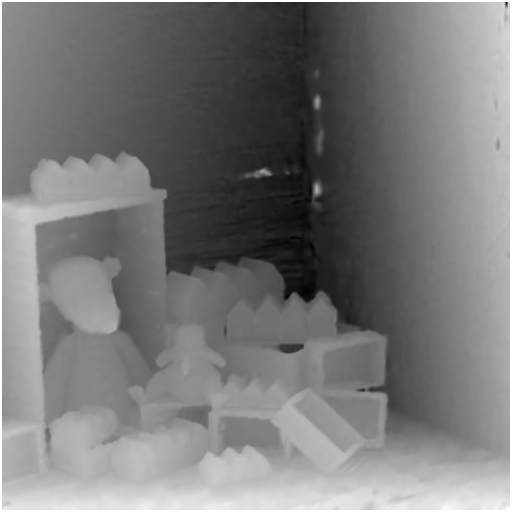} & \includegraphics[height=27.5mm]{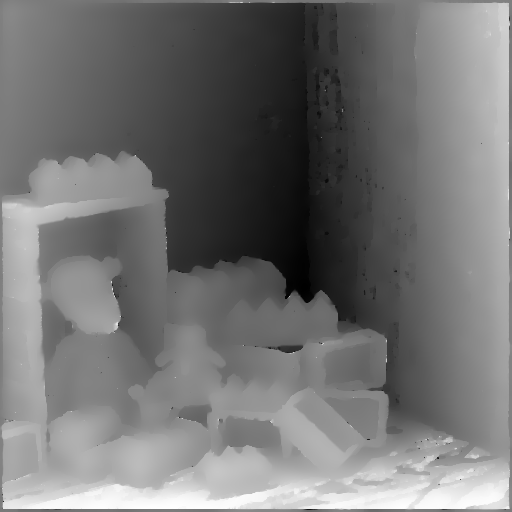} & \includegraphics[height=27.5mm]{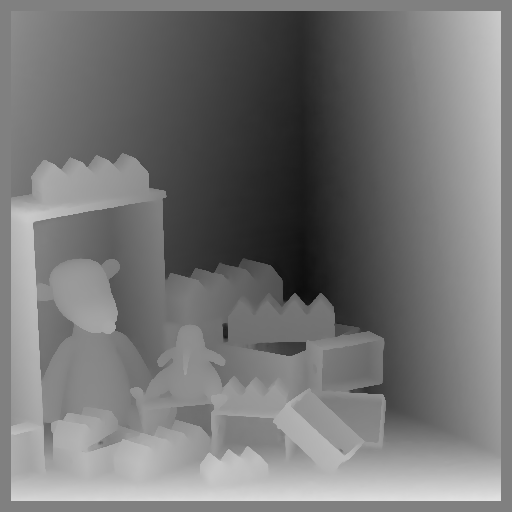}
\\ Original image & Ground truth & BSL \cite{anisimov2018fast} & EPI1 \cite{johannsen2016sparse} & EPI2 \cite{wanner2012globally} & EPINET \cite{shin2018epinet} \\
\includegraphics[height=27.5mm]{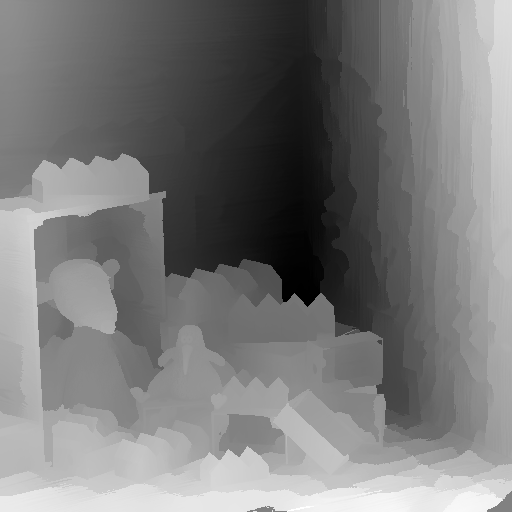} & \includegraphics[height=27.5mm]{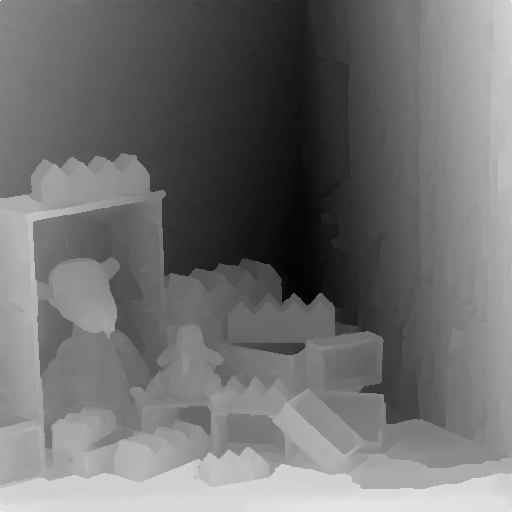} & \includegraphics[height=27.5mm]{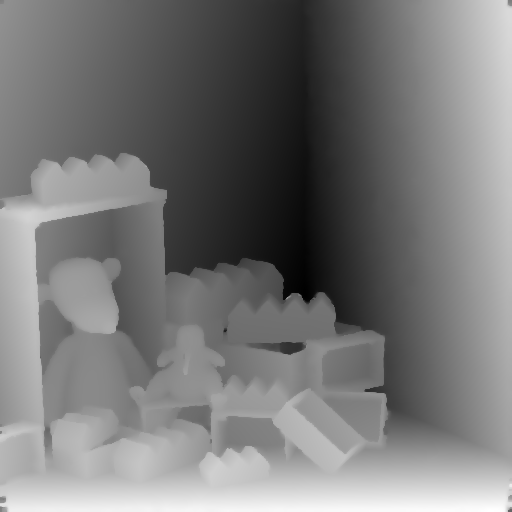} & \includegraphics[height=27.5mm]{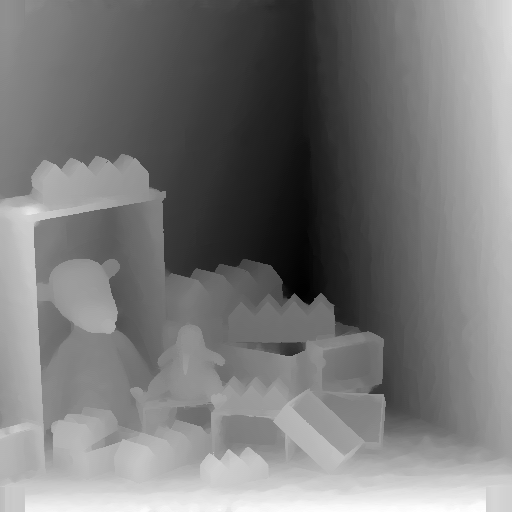} & \includegraphics[height=27.5mm]{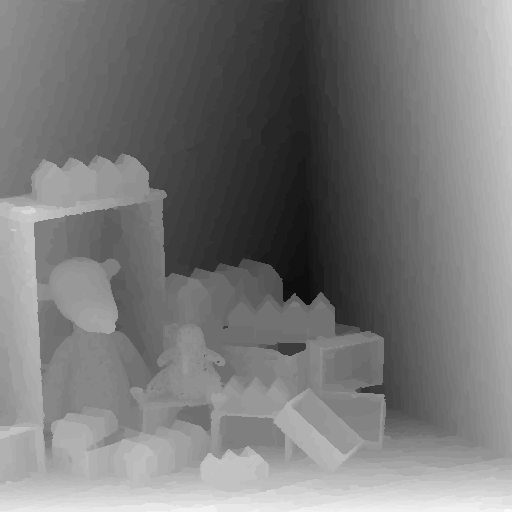} & \includegraphics[height=27.5mm]{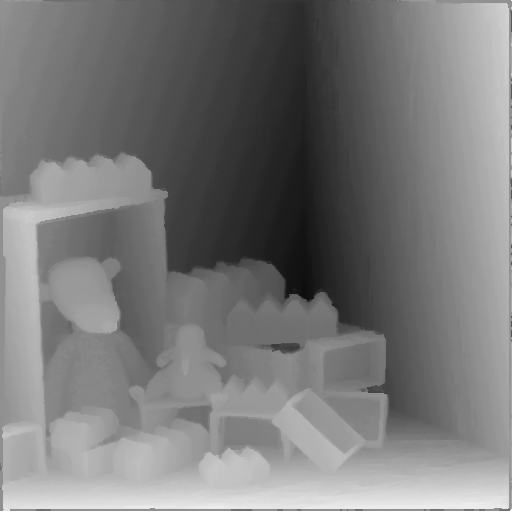} 
\\ LF \cite{jeon2015accurate} & LF\_OCC \cite{wang2015occlusion} & OFSY \cite{strecke2017accurate} & RM3DE \cite{neri2015multi} & RPRF \cite{huang2017robust} & FSL (ours) \\
\end{tabular}
\end{center}
\caption{Qualitative result for "dino" scene from 4D Light Field Benchmark \cite{4DLFB, honauer2016dataset}.}
\label{fig:4dlfb}
\vspace{-2mm}
\end{figure*}
\subsection{Post-processing} \label{final_steps}
Usage of the Census transformation or $\ell_{2}$-norm gives us an opportunity for sub-pixel disparity value estimation.
A popular technique for it is based on parabolic interpolation of cost values.
For each pixel $p$ in interpolated disparity map $D_{n}$ procedure for calculation of the interpolated value can be expressed as follows: 
\begin{align}
\begin{split}
&D_{n}(p) = D_{f}(p)\; +\\ &\frac{C_{s}(p, d-1) - C_{s}(p, d+1)}{2\: (2\: C_{s}(p,d)-C_{s}(p, d-1)-C_{s}(p, d+1))}
\end{split}
\end{align}
Interpolation can only be performed on pixels, in which disparity value $D_{f}(p) \in [D_{L}(p) + 1, D_{H}(p) - 1]$ and $|[D_{L}(p) + 1, D_{H}(p) - 1]| \geqslant 3$. If the disparity value does not satisfy these conditions, then $D_{n}(p) = D_{f}(p)$ 
After this step we apply a median filter to $D_{n}$ in order to remove the impulse noise.
\par
\section{Experiments} \label{experiments}
Our method is evaluated with the 4-dimensional Light Field Benchmark \cite{4DLFB, honauer2016dataset}.
We provide a comparison of the proposed algorithm with the state-of-the-art methods presented in Section \ref{light_field_processing_sota}: BSL \cite{anisimov2018fast}, EPI1 \cite{johannsen2016sparse}, EPI2 \cite{wanner2012globally}, EPINET \cite{shin2018epinet}, LF \cite{jeon2015accurate}, LF\_OCC \cite{wang2015occlusion}, OFSY \cite{strecke2017accurate}, RM3DE \cite{neri2015multi}, RPRF \cite{huang2017robust}. Together with that, qualitative results for real-world EPFL \cite{rerabek2016new} and Middlebury datasets \cite{hirschmuller2007evaluation, scharstein2007learning} are presented.
\par
\subsection{Synthetic dataset}
A dataset with synthetic scenes is provided by 
Honauer \etal \cite{honauer2016dataset} via 4D Light Field Benchmark (4DLFB) \cite{4DLFB}.
Twelve synthetic scenes are used for the comparison; each of them is represented by the 9x9 light field, collected from 8-bit RGB images with 512x512 pixel resolution.
Camera settings and disparity ranges are provided for every scene; high-resolution disparity and depth maps are provided only for categories for training.
\par
\subsection{Evaluation measures} \label{evaluation_measures}
These are several general metrics given by a benchmark.
Fundamental criteria for evaluation of our approach are the estimation of the percentage of pixels, in which absolute difference of the result and ground truth larger than the specified threshold, which is set to 0.07 in our comparison, formulated as the $BadPix$ metric in the mentioned benchmark, together with the runtime of the algorithm.
Algorithms are also compared by the Mean Squared Error ($MSE$) over image pixels.
Corresponding formulas and descriptions are presented by Honauer \etal \cite{honauer2016dataset}, the result for different photo-consistency metrics, which are not covered in this paper, can be found online in the 4D Light Field Benchmark \cite{4DLFB}.
\par
Also, we compare algorithms with $M$-metric \cite{anisimov2018fast}, which is based on the runtime and $BadPix$ metric and formulated as a percentage of correctly computed pixels per second:
\begin{align}
M = \frac{100\% - BadPix}{Runtime}\Big(\frac{\%}{sec.}\Big).
\label{align:metric}
\end{align}
\begin{figure*}
\begin{center}
\centering
\setlength{\tabcolsep}{2pt}
\begin{tabular}{ccccc}
\includegraphics[height=23mm]{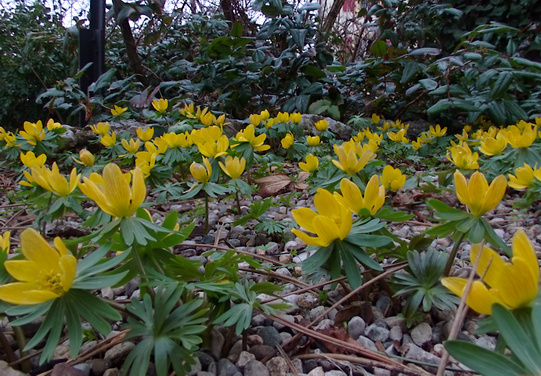} & \includegraphics[height=23mm]{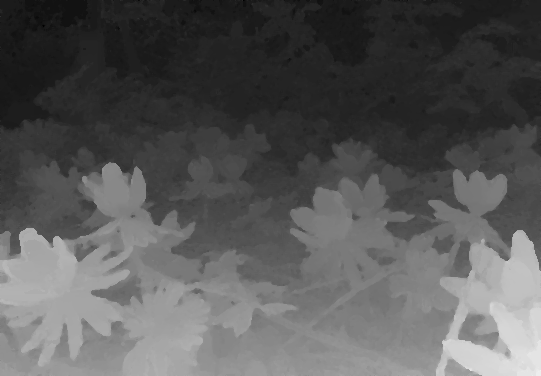} & \includegraphics[height=23mm]{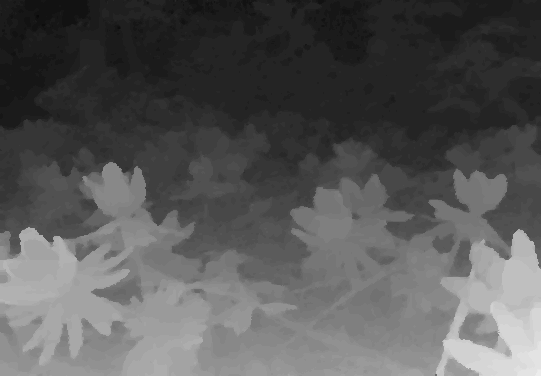} & \includegraphics[height=23mm]{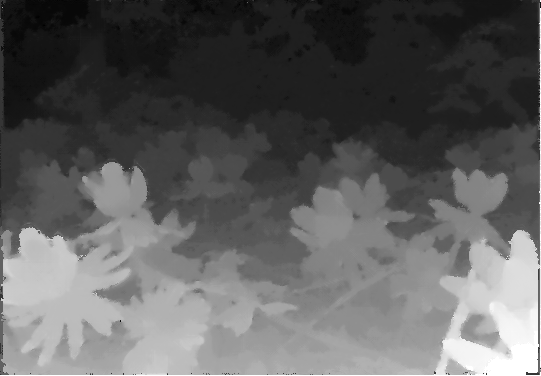} & \includegraphics[height=23mm]{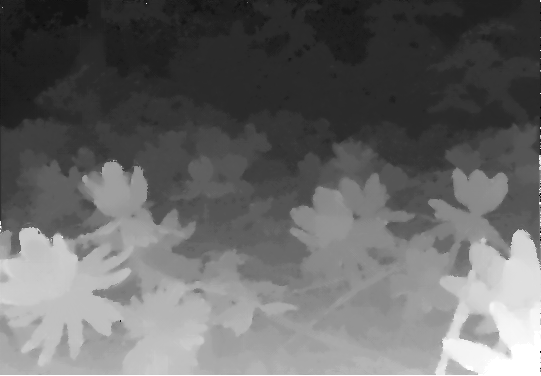}  \\
\includegraphics[height=23mm]{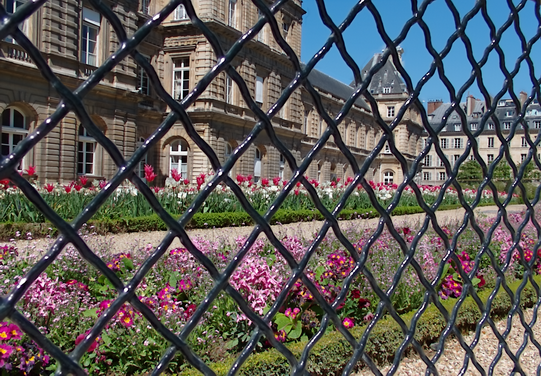} & \includegraphics[height=23mm]{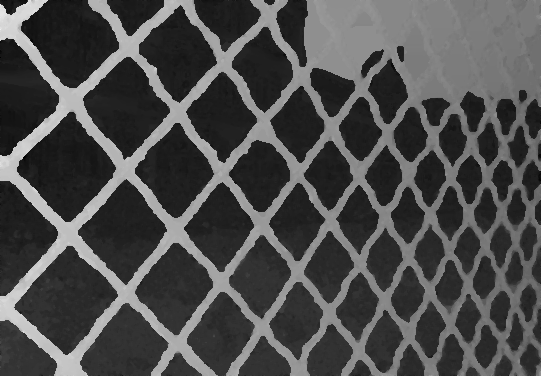} & \includegraphics[height=23mm]{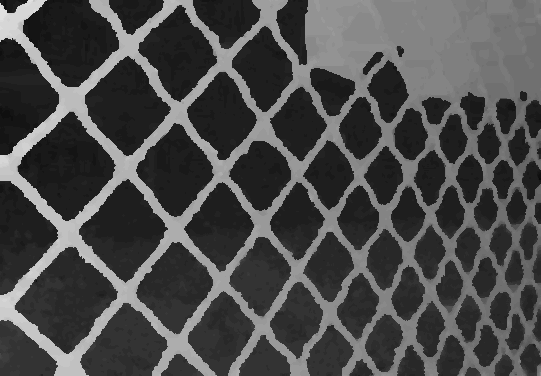} & \includegraphics[height=23mm]{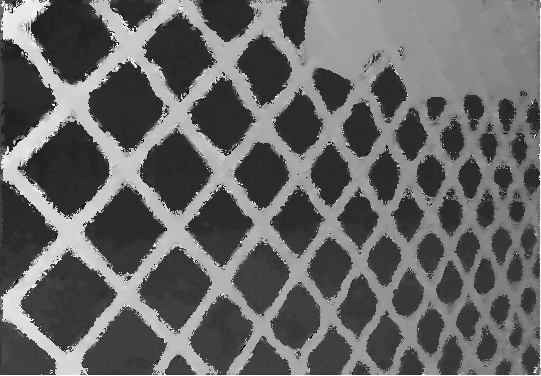} & \includegraphics[height=23mm]{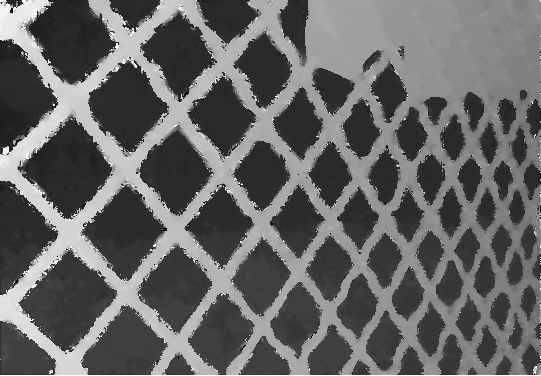} \\
\includegraphics[height=23mm]{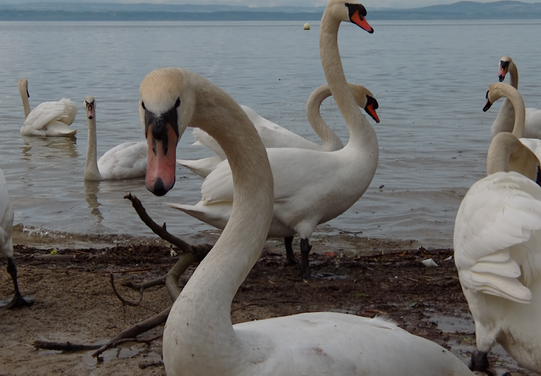} & \includegraphics[height=23mm]{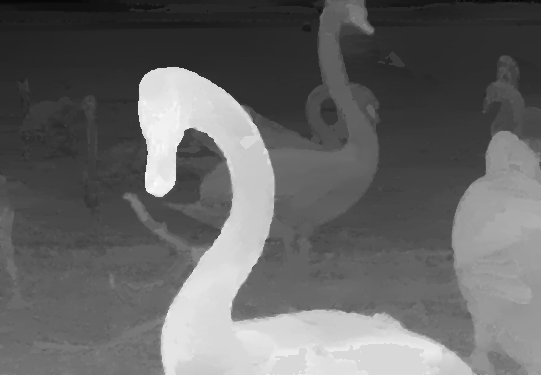} & \includegraphics[height=23mm]{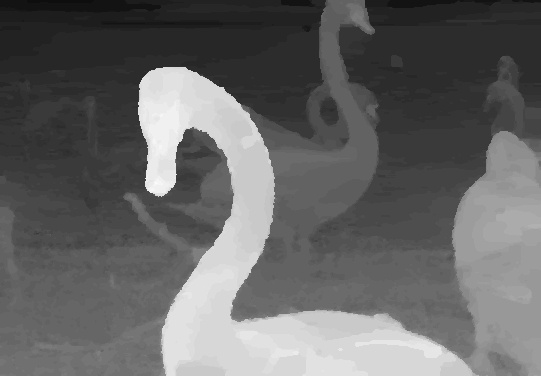} & \includegraphics[height=23mm]{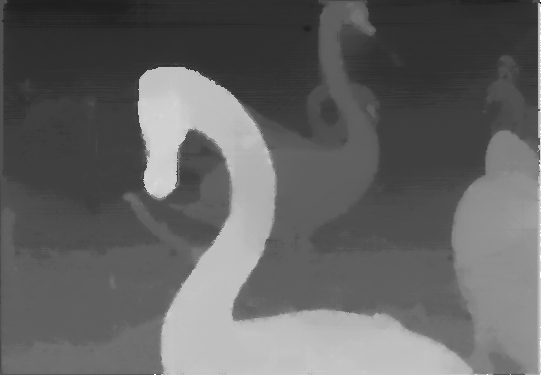} & \includegraphics[height=23mm]{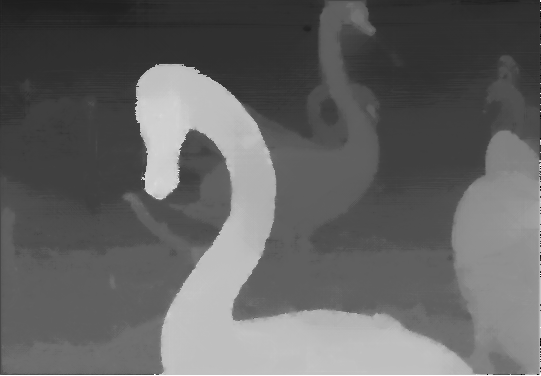} \\
Original image & Lytro depth & RPRF \cite{huang2017robust} & FSL - $\ell_2$ norm & FSL - Census norm \\
\end{tabular}
\end{center}
\caption{Qualitative results for scenes from EPFL dataset \cite{rerabek2016new}}
\label{fig:epfl}
\vspace{-2.5mm}
\end{figure*}

\subsection{Algorithm parameters} \label{algorithm_parameters}
Configuration of our algorithm is adjusted for an optimal result for $BadPix$ metric.
Ranges for depth hypotheses are set accordingly to configuration files of each scene, all other parameters remain same for all scenes.
For the SGM from Section \ref{reducing_of_sampled_disparity_hypotheses} penalty parameters $P\mathit{1}$ and $P\mathit{2}$ are empirically set to 21 and 45 respectively, whereas the penalties for the extended SGM from Section \ref{extended_semi_global_matching} are equal to 17 and 35.
16 traversing directions for SGM are used. 
We use a sparse 7x7 pattern for Census transformation in a configuration from \cite{anisimov2018fast}.
Confidence threshold $\varphi$ is set to 3, and the border penalty $\lambda$ -- to 2.
Window size for median-based filters is equal to 3.
\par
\subsection{Results}
Results of the comparison with metrics from Section \ref{evaluation_measures} are presented in Table \ref{table:4dlfb}.
Visualization of disparity maps for a scene "dino" is presented in Fig.
\ref{fig:4dlfb}; per-scene evaluation together with disparity map results for other scenes can be found online in the 4D Light Field Benchmark \cite{4DLFB} under the FSL acronym.
Comments for the result are provided in Section \ref{discussion}.
\par
\subsection{Real-world scenes} \label{real_world_scene}
Real-world tests were performed with two image similarity measurements for final disparity map estimation. We used an approach, described in Section \ref{extended_semi_global_matching} with  $\ell2$-norm matching cost, and we also modified in to use with Census cost by modifying Eq. \ref{align:S_L2} to:
\begin{align}
\begin{split}
S(u,v,d) = {\sum\limits_{s=1}^n\sum\limits_{t=1}^m}{HD(L_{c}(u, v, \hat s, \hat t), \hat{p}_{c}(u, v, s, t, d))}.
\label{align:S_C}
\end{split}
\end{align}
We provide qualitative results of our algorithm for light field images from EPFL dataset \cite{rerabek2016new}, edited and uploaded by C.-T. Huang for the publication \cite{huang2017robust}.
Light fields consist of 3x3 RGB image with resolution 541x376.
A visualization of the result is presented in Fig.
\ref{fig:epfl}.
On the testing machine in average it took 21 s.
for RPRF \cite{huang2017robust} to generate a result when our approach provides the output in 1.5 s.
\par
\begin{figure*}
\begin{center}
\centering
\setlength{\tabcolsep}{2pt}
\begin{tabular}{ccccc}
\includegraphics[height=27.5mm]{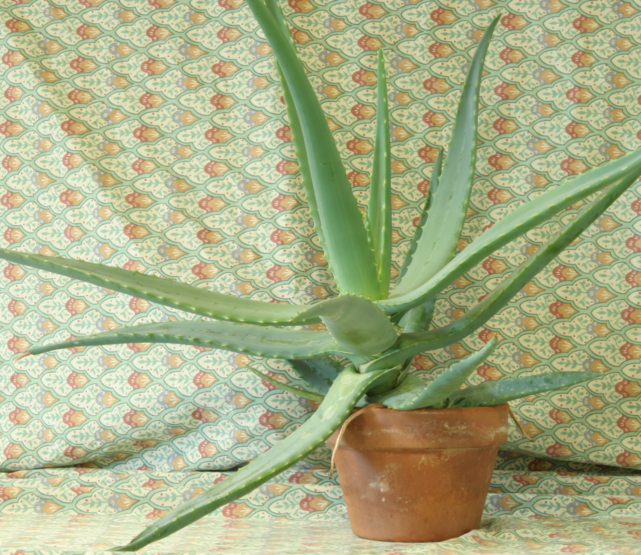} & \includegraphics[height=27.5mm]{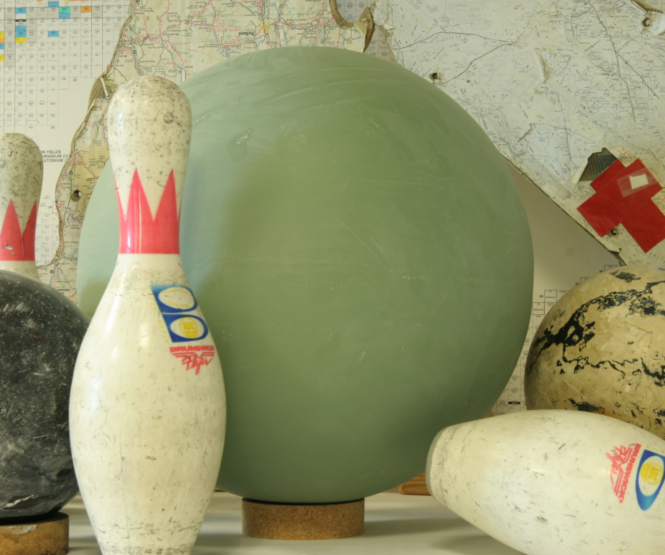} & \includegraphics[height=27.5mm]{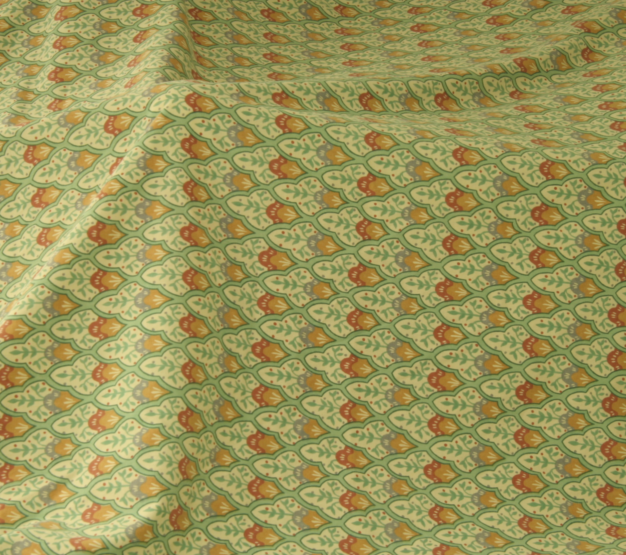} & \includegraphics[height=27.5mm]{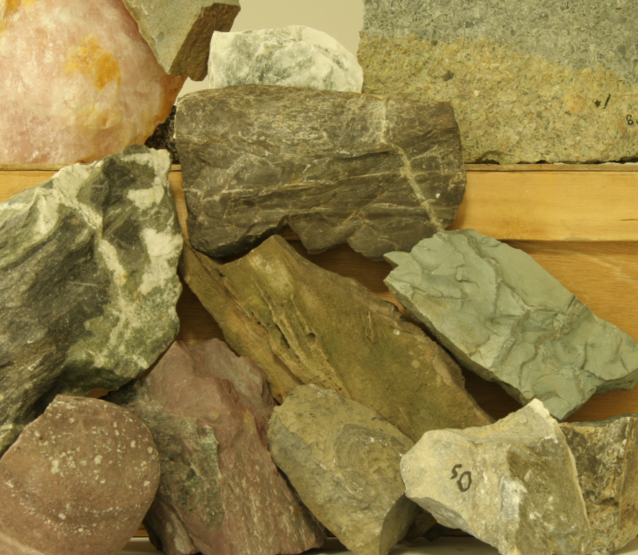} & \includegraphics[height=27.5mm]{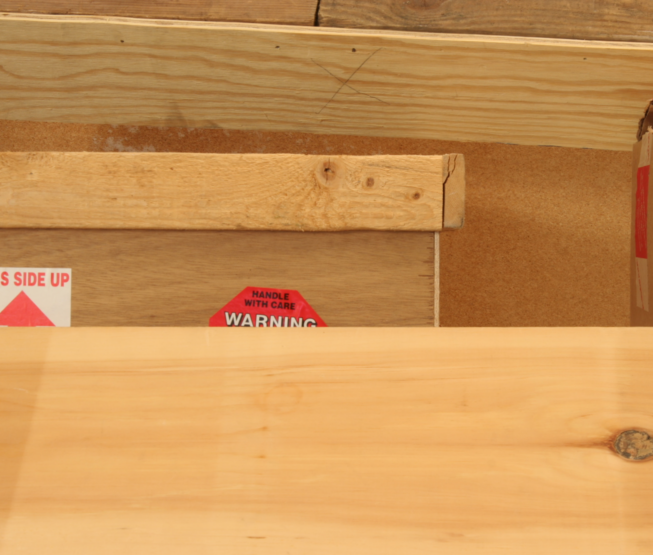} \\
\includegraphics[height=27.5mm]{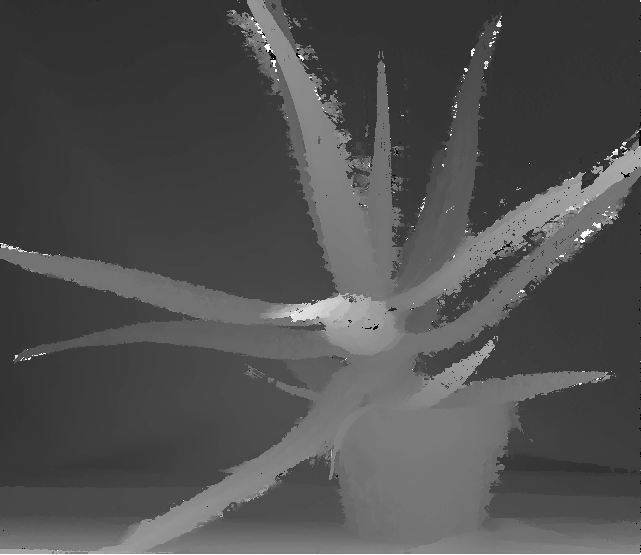} & \includegraphics[height=27.5mm]{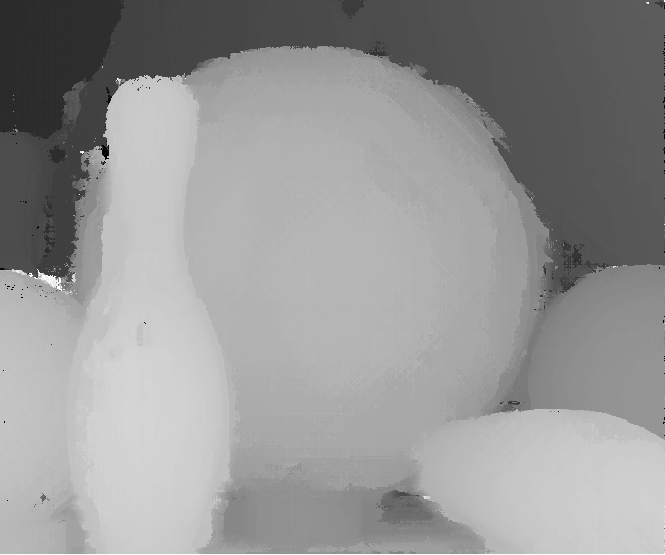} & \includegraphics[height=27.5mm]{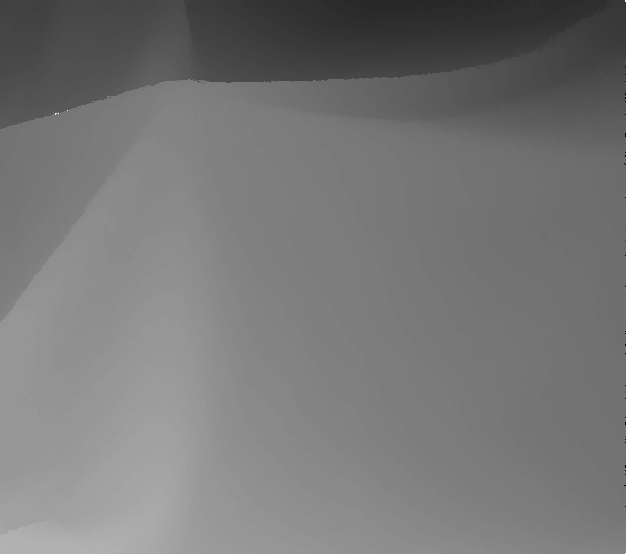} & \includegraphics[height=27.5mm]{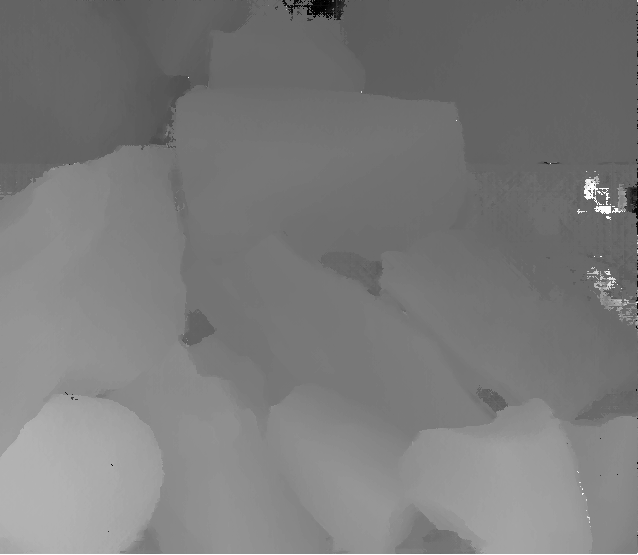} & \includegraphics[height=27.5mm]{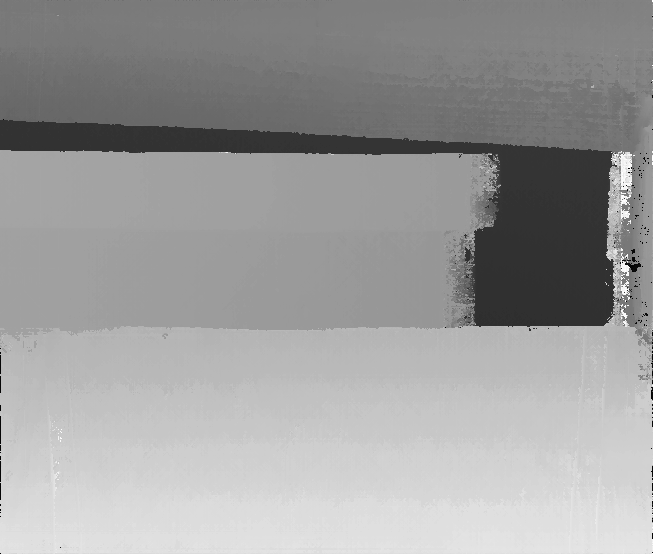} \\
\includegraphics[height=27.5mm]{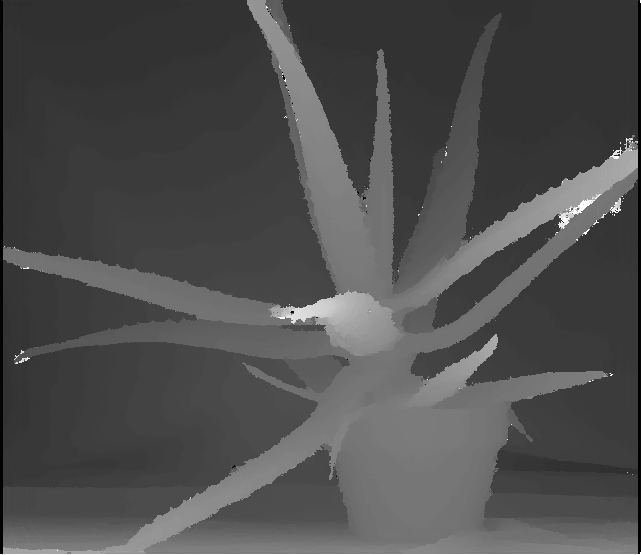} & \includegraphics[height=27.5mm]{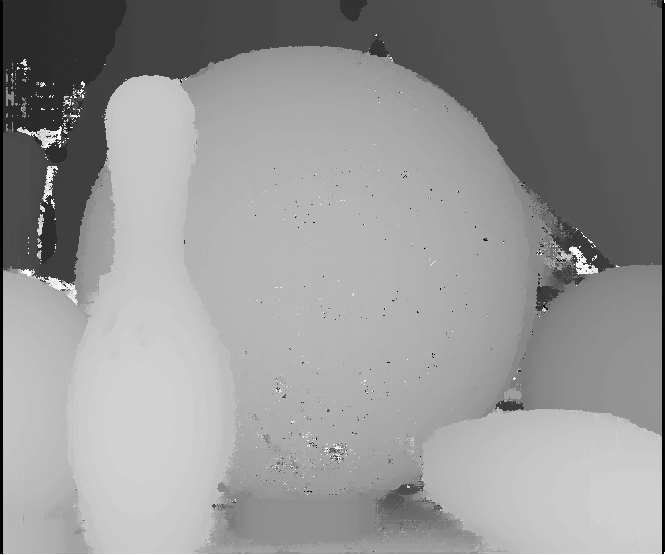} & \includegraphics[height=27.5mm]{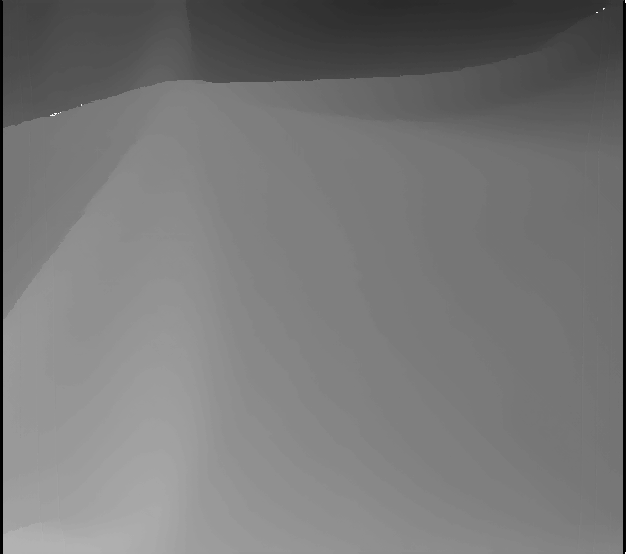} & \includegraphics[height=27.5mm]{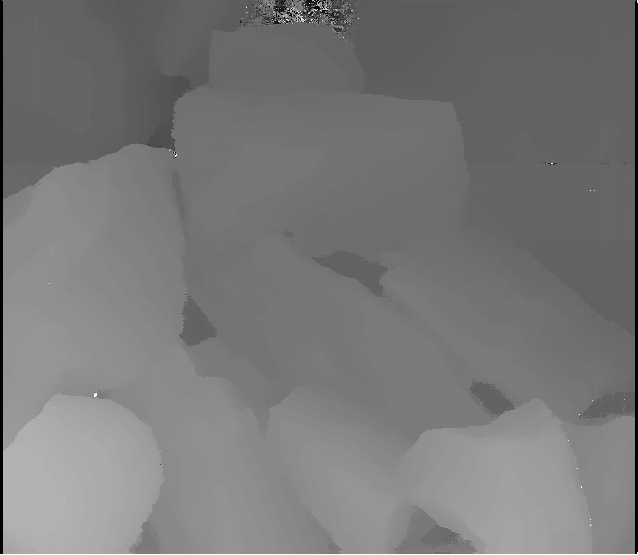} & \includegraphics[height=27.5mm]{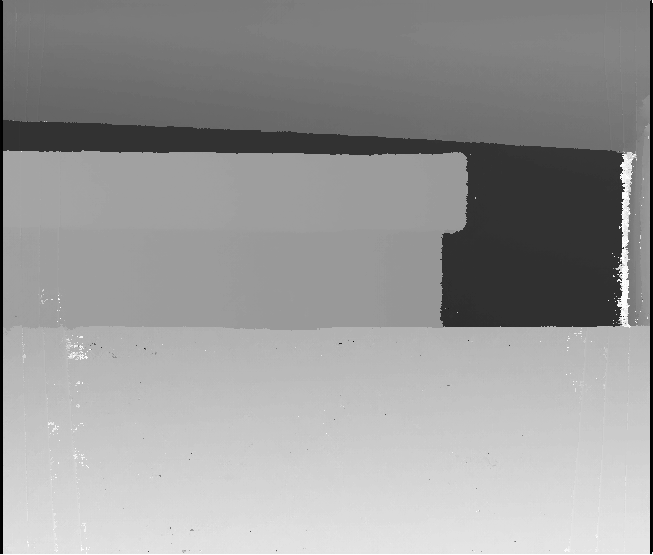} \\
\end{tabular}
\end{center}
\caption{Qualitative results of presented method for Middlebury dataset \cite{hirschmuller2007evaluation, scharstein2007learning}: $\ell_2$-norm in the middle and Census-based in the bottom row}
\label{fig:middlebury}
\vspace{-2.5mm}
\end{figure*}

Also, for additional tests we use 3-dimensional light fields provided by Middlebury 2006 dataset \cite{hirschmuller2007evaluation, scharstein2007learning}.
Each scene is represented by 7 images in a row with a large baseline between them.
Original resolution is 1240-1396x1110 pixels, we used half-sized images for our experiments.
For these images correspondence search is performed in boundary-less manner.
Qualitative result for some scenes with different image similarity measurements is visualized in Fig. \ref{fig:middlebury}.
We do not provide ground truth images for this comparison, because they are available only for two bordering views on the light fields, and we compute our disparity map with respect to the central light field image.
Processing of these scenes took 8 s. on average.
For both datasets, penalty parameters for SGM are set to 30 and 120, other algorithm parameters remain same and described in \ref{algorithm_parameters}.
\par
\subsection{Discussion} \label{discussion}
The results of our algorithm are expectedly consistent with results from BSL \cite{anisimov2018fast}.
Quantitatively, the average values of $BadPix$ and MSE are improved, which can be explained as an effect of applying SGM to the generated costs.
In general, our algorithm produces average quality results within other algorithms.
Qualitatively we can see an improvement of the subjective sharpness level compared to BSL.
Comparison based on photo-realistic metrics proposed by Honauer \etal \cite{honauer2016dataset} also shows a slight improvement of the result.
However, a visual issue with step effect on the disparity map is recognized.
Further experiments are required to determine exactly how this problem can be eliminated.
We believe that this effect can be partially compensated by proper adjustment of penalty parameters for the extended SGM or with more sophisticated post-processing filtration.
\par
Currently, we achieve the best result for runtime and the second best for M-metric (\ref{align:metric}).
The approach EPINET \cite{shin2018epinet} can be considered as top-of-the-line, providing good results in terms of depth quality together with the satisfying runtime.
However, their algorithm is performed on the high-end GPU.
Such heavy computational resources are not required by our approach, which utilizes a central processing unit (CPU) without specifically employed thread parallelism. The exact configuration is explained in Section \ref{environment}.
There is a field of improvement for this algorithm with SIMD-instructions and exploitation of parallelism \eg for independent of each other traversing directions.
\par
Borders from the initial depth map help to reduce the number of sampled hypotheses by further processing in a range from 50\% (real-world scenes) up to 97\% (synthetic scenes).
Runtime of the correspondence search in the light field space is affected proportionally.
\par
A possible gain of the number of bordered pixels can be achieved by the improvement of the matching algorithm for initial disparity map estimation.
For instance, it can be done with different configurations of Census window \eg by using adaptive Census window based on gradients.
Real-world tests show that utilization of $\ell_2$-norm for global correspondence search cannot be considered as effective in some cases.
$\ell_{2}$-norm works better for the case of synthetic images and real-world scenes with narrow baseline in light field images, whereas Census-based estimation provides a better result for other real-world cases.
Configuration with Census-based matching cost tested with Middlebury dataset images (Fig. \ref{fig:middlebury}) revealed a fact that computation of the initial disparity map and utilization of it for correspondence search with the mentioned similarity measurement shows worse runtime compare to the configuration with initialization.
In our opinion, it happens mainly because of a large number of depth hypotheses in this dataset.
The initial map was computed for EPFL dataset for both similarity configurations, and the time difference was insignificant.
\par
We assume that decision for the selection of proper matching cost and the boundaries configuration should depend on the image size, the maximum possible displacement between two boundary light field views and quality of the image by itself (\eg it should not be noisy).
Without boundaries, Census-based configuration has somewhat similar to the method of Tomioka \etal \cite{tomioka2017depth} in terms of aggregation principles.
\par
During the tests with real-world images, we came up with an idea of adjustment for border threshold $\lambda$ dependent on the initial depth value, so that for small disparities this threshold is higher rather than for higher values.
Such a strategy can help to determine the more accurate values for the farther objects, where the pixel discontinuity can be a crucial factor for the right depth estimation.
\par
Additional limitations have been observed during these tests.
On Fig.
\ref{fig:epfl} the visual problem with grids in the middle scene can be noticed.
Also, we were not able to generate proper interpolation for a bottom scene in Fig.
\ref{fig:epfl}, it could be a problem related to the use of a small number of images for reconstruction.
Although bordering of depth values with the initial map helps to reduce depth mismatching noise, some of the wrongly calculated pixels still can "survive" this filtering, which is visible in images on Fig.
\ref{fig:middlebury}.
These mistakes appear either in the areas marked previously as non-consistent or on the object edges.
Our algorithm fails with depth estimation on image boundaries for scenes with a relatively large distance between images.
The explanation of this problem is related to our selection of the central light field image as a reference view.
In this case search for a matching pixel from image boundary fails since there is no match in most of the images and therefore our algorithm can not aggregate enough depth score for the correct value.
A solution for this problem was proposed by Kim \etal \cite{kim2013scene}, where the position of reference images changes over time and cost aggregation is performed from the new position.
However, for the presented method such an improvement can greatly increase the running time, which is in contradiction with our main objective.
\subsection{Environment} \label{environment}
Hardware configuration for depth processing includes CPU Intel Xeon E3-1245 V2 @ 3.40 GHz, forced to work in single-thread mode.
Our algorithm is implemented in C and compiled using GCC v.7.3.1 with /O3 option.
\par
\section{Conclusion} \label{conclusion}
In this paper, we proposed a fast depth estimation method from light fields by extending an efficient stereo matching algorithm with a methodology of enlargement to multi-view sampling by the correspondence search in light field space.
The evaluation against state-of-the-art methods showed that our algorithm produces comparable depth map results, which have been proven by different quantitative metrics.
In contrast to many state-of-the-art approaches, our proposed method produces depth maps in a relatively small amount of time.
As a result, our method provides almost the best result in terms of computed pixels per unit of time.
In addition, our approach works for different configurations of 3-dimensional and 4-dimensional synthetic and real-world light field images.
Further work will investigate additional real-time improvements, \eg by utilizing a pyramidal scheme with downscaled images together with hardware-specified optimizations, such as parallelism on multi-core CPU or GPU.
\par
\section*{Acknowledgments}
This work was partially funded by the Federal Ministry of Education and Research (Germany) in the context of the project DAKARA (13N14318).
The authors are grateful to Kiran Varanasi and Jonathan Wray for the provided support.
\bibliographystyle{IEEEtran}
\bibliography{egbib}

\end{document}